# Random Shuffling and Resets for the Non-stationary Stochastic Bandit Problem

Robin Allesiardo [1,2] · Raphaël Féraud [1] · Odalric-Ambrym Maillard [2].

[1] Orange Labs

[2] Team TAO, CNRS - INRIA - LRI

September 7, 2016

### Abstract

We consider a non-stationary formulation of the stochastic multi-armed bandit where the rewards are no longer assumed to be identically distributed. For the best-arm identification task, we introduce a version of SUCCESSIVE ELIMINATION based on random shuffling of the $K$ arms. We prove that under a novel and mild assumption on the mean gap $\Delta$, this simple but powerful modification achieves the same guarantees in term of sample complexity and cumulative regret than its original version, but in a much wider class of problems, as it is not anymore constrained to stationary distributions. We also show that the original SUCCESSIVE ELIMINATION fails to have controlled regret in this more general scenario, thus showing the benefit of shuffling. We then remove our mild assumption and adapt the algorithm to the best-arm identification task with switching arms. We adapt the definition of the sample complexity for that case and prove that, against an optimal policy with $N-1$ switches of the optimal arm, this new algorithm achieves an expected sample complexity of $O(\Delta^{-2}\sqrt{NK\delta^{-1}\log(K\delta^{-1})})$, where $\delta$ is the probability of failure of the algorithm, and an expected cumulative regret of $O(\Delta^{-1}\sqrt{NTK\log(TK)})$ after $T$ time steps.

## 1 Introduction

The theoretical framework of the multi-armed bandit problem formalizes the fundamental exploration/exploitation dilemma that appears in decision making problems facing partial information. At a high level, a set of $K$ arms is available to a player. At each turn, she has to choose one arm and receives a reward corresponding to the played arm, without knowing what would have been the received reward had she played another arm. The player faces the dilemma of *exploring*, that is playing an arm whose mean reward is loosely estimated in order to build a better estimate or *exploiting*, that is playing a seemingly best arm based on current mean estimates in order to maximize her cumulative reward. The accuracy of the player policy at time horizon $T$ is typically measured in terms of *sample complexity* or of *regret*. The sample complexity is the number of plays required to find an approximation of the best arm with high probability.

In that case, the player can stop playing after identifying this arm. The regret is the difference between the cumulative rewards of the player and the one that could be acquired by a policy assumed to be optimal.

The **stochastic** multi-armed bandit problem assumes the rewards to be generated independently from stochastic distribution associated with each arm. Stochastic algorithms usually assume distributions to be constant over time like with the Thompson Sampling (TS) [13], UCB [2] or Successive Elimination (SE) [4]. Under this assumption of *stationarity*, TS and UCB achieve optimal upper-bounds on the cumulative regret with logarithmic dependencies on $T$. SE algorithm achieves near optimal sample complexity.

In the **adversarial** multi-armed bandit problem, rewards are chosen by an adversary. This formulation can model any form of non-stationarity. The EXP3 algorithm [3, 11] achieves an optimal regret of $O(\sqrt{T})$ against an oblivious opponent that chooses rewards before the beginning of the game, with respect to the best policy that pulls the same arm over the totality of the game. This weakness is partially overcomed by EXP3.S [3], a variant of EXP3, that forgets the past adding at each time step a proportion of the mean gain and achieves controlled regret with respect to policies that allow arm switches during the run.

The **switching bandit** problem introduces non-stationarity within the *stochastic* bandit problem by allowing means to change at some time-steps. *Discounted* UCB [10] and *sliding-window* UCB [6] are adaptations of UCB to the switching bandit problem and achieve a regret bound of $O(\sqrt{MT\log T})$, where $M - 1$ is the number of distribution changes. EXP3.R [1] combines EXP3 with a switch detector based on Hoeffding inequality [8] in order to detect switches of best arm. This detector does not require mean rewards to be constant on the tested interval. This use of the Hoeffding inequality on nonidentical distributions is extented to SE in this paper to allow the algorithm to achieve efficient theoretical guarantees on non-stationary distributions of rewards. EXP3.R achieves a regret bound of $O(N\sqrt{T\log T})$, where $N - 1$ is the number of times when the best arm changes. As this is always smaller (and potentially much smaller) than the number of distribution changes ($N \leq M$), EXP3.R appears to be a strong competitor. It is also worth citing META-EVE [7] that associates UCB with a mean change detector, resetting the algorithm when a change is detected. While no analysis is provided, it has demonstrated strong empirical performances.

**Stochastic and Adversarial**. Several variants combining stochastic and adversarial rewards have been proposed by Seldin & Slivkins [12]. For instance, in the setting with *contaminated rewards*, rewards are mainly drawn from stationary distributions except for a minority of rewards of means chosen in advance by an adversary. In order to guarantee their proposed algorithm EXP3++ achieves logarithmic guarantees, the adversary is constrained in the sense it cannot lowered the gap between arms more than a factor $1/2$. They also proposed another variant called *adversarial with gap* [12] which assumes the existence of a round after which an arm persists to be the best. This work is motivated by the desire to create generic algorithms able to perform bandit tasks with various reward types, stationary, adversary or mainly stationary. However, despite achieving good performances on a wide range of problems, each one needs a specific parametrization (i.e. an instance of EXP3++ parametrized for stationnary rewards may not perform well if rewards are choosen by an adversary).

**Our contribution**. We consider a formalization of the bandit problem where rewards are drawn from stochastic distributions of arbitrary means defined before the beginning of the game. In this non-stationary stochastic bandit context, our first contribution is to introduce and analyze a randomized version of SUCCESSIVE ELIMINATION. We show that the seemingly minor modification – a randomized round-robin procedure – leads to a dramatic improvement of the performance over the original SUCCESSIVE ELIMINATION algorithm. We show for instance in Theorem 1 and Corollary 1 that the proposed SUCCESSIVE ELIMINATION WITH RANDOMIZED ROUND-ROBIN (SER3) algorithm achieves a controlled sample complexity and cumulative regret in situations where SUCCESSIVE ELIMINATION may even suffers from a linear regret in the time horizon $T$, as supported by numerical experiments (see Section 5). Our second contribution is to identify a notion of gap that generalizes the gap from stochastic bandits to the non-stationary case, and derive *gap-dependent* (also known as problem-dependent) sample complexity and regret bounds, instead of the more classical and less informative *problem-free* bounds.

This paper is organized as follows. Section 3 starts with the case of non-stationary stochastic bandits when the best arm does not change during the game. We show that the sample complexity of SER3 is controlled as $O(\frac{K}{\Delta^2} \log \frac{K}{\delta \Delta})$ where $\Delta$ generalizes the usual notion of stochastic gap from the stationary case. This comes under a mild assumption (see assumption 1). In Section 4, we then extend the setting to a full-blown switching problem when the best arm is allowed to change over time. We naturally extend the definition of sample complexity to allow the optimal arm to change over time (see Definition 1), and introduce the RANDOMIZED SUCCESSIVE ELIMINATION WITH RESETS (SER4) algorithm. We show in Theorem 2 and Corollary 2 that this algorithm achieves an expected sample complexity of $O(\sqrt{NK \log(K/\delta)}/(\delta \Delta^2))$ where $N - 1$ is the number of arm switches assumed to be known. Corollary 3 extend this result to the expected cumulative regret with an upper-bound of $O(\Delta^{-1}\sqrt{NTK \log(TK)})$. These results do not assume any constraint on the behavior of the stochastic distributions and provide a gap-dependent performance bound in the challenging setting of non-stationary stochastic bandits competing against switching best arms. We finally illustrate the proposed approaches on synthetic problems in Section 5.

## 2 Setting

We consider a generalization of the adversarial setting where the adversary chooses before the beginning of the game a sequence of *distributions* instead of directly choosing a sequence of rewards. This generalizes the setting since choosing arbitrarily a reward $y_k(t)$ is equivalent to drawing this reward from a distribution of mean $y_k(t)$ and a variance of zero.

**The problem.** Let $[K] = 1, ..., K$ be a set of $K$ arms. The reward $y_{k_t}(t) \in [0, 1]$ obtained by the player after playing the arm $k_t$ is drawn from a distribution of mean $\mu_{k_t}(t) \in [0, 1]$. The instantaneous gap between arms $k$ and $k'$ at time $t$ is:

$$\Delta_{k,k'}(t) \stackrel{\text{def}}{=} \mu_k(t) - \mu_{k'}(t). \tag{1}$$

Let $k^*(t)$ be the arm played by the optimal policy at time $t$.

**The notion of sample complexity.** In the literature [9], the sample-complexity of an algorithm is the number of samples needed by this algorithm to find a policy achieving a specific level of performance with high probability. We denote $\delta \in (0, 0.5]$ the probability of failure. For instance, for the best arm identification in the stochastic stationary bandit (that is when $\forall k \forall t, \mu_k(t) = \mu_k(t+1)$ and $k^*(t) = k^*(t+1)$), the sample complexity is the number of sample needed to find, with a probability at least $1 - \delta$, an arm $k^*$ of mean reward $\max_{k \in [K]} \mu^k$. In section 3.1 we define the best arm in the stochastic non-stationary bandit problem without switches and in section 4 we propose a new definition of the sample complexity for the best arm identification problem with switches.

Analysis in sample complexity are useful for building hierarchical models of contextual bandits in a greedy way [5], reducing the exploration space.

**The notion of regret.** We define the pseudo cumulative regret as the difference of mean rewards between the arms chosen by the optimal policy and those chosen by the player:

$$\sum_{t=1}^{T} \mu_{k^*(t)}(t) - \mu_{k_t}(t). \tag{2}$$

Usually, in the stochastic bandit setting, the distributions of rewards are stationary and the instantaneous gap $\Delta_{k,k'}(t) = \mu_k(t) - \mu_{k'}(t)$ is the same for all the time-steps.

## 3 Best Arm Identification in Non-stationary Stochastic Bandits.

In this section, we present the algorithm SUCCESSIVE ELIMINATION WITH RANDOMIZED ROUND-ROBIN (SER3, see algorithm 1), a randomized version of SUCCESSIVE ELIMINATION which tackles the best arm identification problem when rewards are non-stationary.

### 3.1 A modified Successive Elimination algorithm

We elaborate on several notions required to understand the behavior of the algorithm and to relax the constrain of stationarity.

**The elimination mechanism** The elimination mechanism was introduced by SUCCESSIVE ELIMINATION [4]. Estimators of the rewards are built by sequentially sampling the arms. After $\tau_{\min}$ turns of round-robin, the elimination mechanism starts to occur. A lower-bound of the reward of the best empirical arm is computed and compared with an upper-bound of the reward of all other arms. If the lower-bound is higher than one of the upper-bounds, then the associated arm is eliminated and stop being considered by the algorithm. Processes of sampling and elimination are repeated until the elimination of all arms except one.

---

**Algorithm 1** SUCCESSIVE ELIMINATION WITH RANDOMIZED ROUND-ROBIN (SER3)

---

   **input:** $\delta \in (0, 0.5]$, $\epsilon \in [0, 1)$, $\tau_{\min} = \log \frac{K}{\delta}$
   **output:** an $\epsilon$-approximation of the best arm
   $S_1 = [K]$, $\forall k$, $\hat{\mu}_k(0) = 0$, $t = 1$, $\tau = 1$
   **While** $|S_\tau| > 1$
     Shuffle $S_\tau$
     **For** each $k \in S_\tau$ **do**
       Play $k$
       $\hat{\mu}_k(\tau) = \frac{\tau-1}{\tau}\hat{\mu}_k(\tau-1) + \frac{y_k(t)}{\tau}$
       $t = t + 1$
     **End for**
     $k_{\max} = \arg\max_{k \in S} \hat{\mu}_k(\tau)$
     **If** $\tau \geq \tau_{\min}$
       Remove from $S_{\tau+1}$ all $k$ such as:

$$\hat{\mu}_{\max}(\tau) - \hat{\mu}_k(\tau) + \epsilon \geq \sqrt{\frac{2}{\tau}\log\left(\frac{4K\tau^2}{\delta}\right)}$$

     **End if**
     $\tau = \tau + 1$
   **End while**

---

**Hoeffding inequality.** SUCCESSIVE ELIMINATION assumes that the rewards are drawn from stochastic distributions that are identical over time (rewards are identically distributed). However, the Hoeffding inequality used by this algorithm does not require stationarity and only requires independence. We remember the Hoeffding inequality:

**Lemma 1** (Hoeffding inequality [8]). *If $X_1, X_2, ..., X_\tau$ are $\tau$ independent random variables and $0 \leq X_i \leq 1$ for all $(i = 1, 2, ..., \tau)$, then for $\epsilon_\tau > 0$*

$$P\left(\left|\sum_{i=1}^{t}\frac{X_i}{\tau} - \mathbb{E}\left[\sum_{i=1}^{t}\frac{X_i}{\tau}\right]\right| \geq \epsilon_\tau\right) \leq 2\exp\left(-2\epsilon_\tau^2 \tau\right).$$

Thus we can use this inequality to calculate confidence bounds of empirical means computed with rewards drawn from non identical distributions.

**Randomization of the Round-Robin.** We illustrate the need of randomization with an example tricking a deterministic algorithm (see figure 1).

| $\mu_k(t)$ | $t=1$ | $t=2$ | $t=3$ | $t=4$ | $t=5$ | $t=6$ |
|---|---|---|---|---|---|---|
| $k=1$ | 0.6 | 1 | 0.6 | 1 | 0.6 | 1 |
| $k=2$ | 0.4 | 0.8 | 0.4 | 0.8 | 0.4 | 0.8 |

Figure 1: A sequence of mean rewards tricking a deterministic bandit algorithm.

The best arm seem to be $k = 1$ as $\mu_1(t)$ is greater than $\mu_2(t)$ at every time-step $t$. However, by sampling the arms with a deterministic policy playing sequentially $k = 1$ and then $k = 2$, after $t = 6$ the algorithm has only sampled rewards from a distribution of mean 0.6 for $k = 1$ and of mean 0.8 for $k = 2$. After enough time following this pattern, an elimination algorithm will eliminate the first arm. Our algorithm SER3 adds a shuffling of the arm set after each round-robin cycle to SUCCESSIVE ELIMINATION and avoids this behavior.

**Uniqueness of the best arm.** The best arm identification task assumes a criteria identifying the best arm without ambiguity. We define the **optimal arm** as:

$$k^* = \arg\max_{k \in [K]} \sum_{t=1}^{T} \mu_k(t). \quad (3)$$

As an efficient algorithm will find the best arm before the end of the run, we use assumption 1 to ensure its uniqueness at every time-step. First, we define some notations. A run of SER3 is a succession of round-robin. The set $[\tau] = \{(t_1, |S_1|), ..., (t_\tau, |S_\tau|)\}$ is a realization of SER3 and $t_i$ is the timestep when the round-robin $i^{\text{th}}$ of size $|S_i|$ starts ($t_i = 1 + \sum_{j=1}^{i-1} |S_j|$). As arms are only eliminated, $|S_i| \geq |S_{i+1}|$. We denote $\mathbb{T}(\tau)$ the set containing all possible realizations of $\tau$ round-robin steps. Now, we can introduce assumption 1 that ensures the best arm is the same at any time-step.

**Assumption 1** (Positive mean-gap). *For any $k \in [K] - \{k^*\}$ and any $[\tau] \in \mathbb{T}(\tau)$ with $\tau \geq \tau_{\min}$, we have:*

$$\Delta_k^*([\tau]) = \frac{1}{\tau} \sum_{i=1}^{\tau} \sum_{j=t_i}^{t_i+|S_i|-1} \frac{\Delta_{k^*,k}(j)}{|S_i|} > 0. \quad (4)$$

Assumption 1 is quite weak (see e.g. figure 2(b)) and can tolerate a large noise when $\tau$ is high. As the optimal arm must distinguish itself from others, instantaneous gaps are more constrained at the beginning of the game. It is quite similar to the assumption used by Seldin & Slivkins [12] to be able to achieve logarithmic expected regret on *moderately contaminated rewards*, i.e., the adversary does not lower the averaged gap too much. Another analogy can be done with the *adversarial with gap* setting [12], $\tau_{\min}$ representing the time needed for the optimal arm to accumulate enough rewards and to distinguish itself from the suboptimal arms.

Figure 2(a) illustrates assumption 1. In this example the mean of the optimal arm $k^*$ is lower than the second one on time-steps $t \in \{5, 6, 7\}$. Thus even if the instantaneous gap

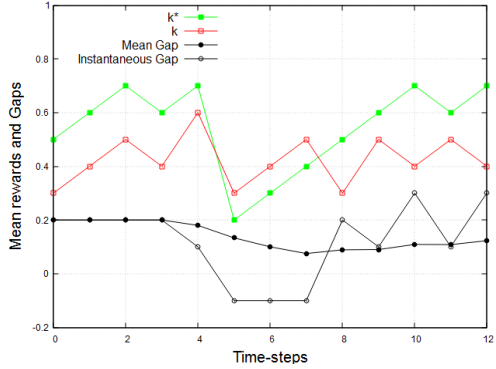

(a) Assumption 1 is satisfied as the mean gap remains positive.

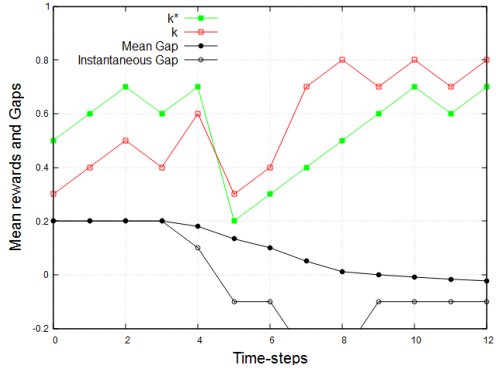

(b) Assumption 1 is not satisfied. This sequence involves a best arm switch as the mean gap become non positive.

Figure 2: Two examples of sequence of mean rewards.

is negative during these time-steps, the mean gap $\Delta_k^*([\tau])$ stays positive. The parameter $\tau_{\min}$ protects the algorithm from local noise at the initialization of the algorithm. In order to ease the reading of the results, we here assume $\tau_{\min} = \log \frac{K}{\delta}$.

Assumption 1 can be seen as a sanity-check assumption ensuring that the best-arm identification problem indeed makes sense. In section 4, we consider the more general switching bandit problem. In this case, assumption 1 may not be verified (see figure 2(b)), and is naturally extended by dividing the game in segments wherein assumption 1 is satisfied.

## 3.2 Analysis

The sample-complexity is the number of observations needed to find an $\epsilon$-optimal arm with high probability. All theoretical results are provided for $\epsilon = 0$ and therefore accept only $k^*$ as the optimal arm. *All proofs are provided in the appendix.*

**Theorem 1.** *For $K \geq 2$, $\delta \in (0, 0.5]$, and $\tau_{\min} = \log \frac{K}{\delta}$, the sample-complexity of* SER3 *is upper bounded by:*
$$O\left(\frac{K}{\Delta^2} \log(\frac{K}{\delta \Delta})\right)$$
*where* $\Delta = \min_{[\tau],k} \frac{1}{\tau} \sum_{i=1}^{\tau} \sum_{t=t_i}^{t_i+|S_i|-1} \frac{\Delta_{k^*,k}(t)}{|S_i|}$.

Guarantee on the sample complexity can be transposed in guarantee on the cumulative regret. In that case, when only one arm remains in the set, the player continues to play this last arm until the end of the game.

**Corollary 1.** *For $K \geq 2$, and $\delta = 1/T$, and $\tau_{\min} = \log(KT)$, the expected cumulative regret of* SER3 *is upper bounded by:*
$$\min\left(O\left(\frac{K-1}{\Delta} \log(\frac{KT}{\Delta})\right), O\left(\sqrt{TK \log \frac{T}{K}}\right)\right)$$

These guarantees are the same as the original SUCCESSIVE ELIMINATION performed with a deterministic round-robin on arms with stationary rewards. Indeed, when reward distributions are stationary, we have for all $t$ and all $[\tau]$:
$$\frac{1}{\tau} \sum_{i=1}^{\tau} \sum_{t=t_i}^{t_i+|S_i|-1} \frac{\Delta_{k^*,k}(t)}{|S_i|} = \Delta_{k^*,k}(t) = \Delta_{k^*,k}(t+1). \tag{5}$$

However, in a non-stationary environment satisfying assumption 1 SUCCESSIVE ELIMINATION will eliminate the optimal arm if the adversary knows the order of its round-robin before the beginning of the run and exploits this knowledge against the learner, thus resulting in a linear cumulative regret.

**Remark:** These logarithmic guarantees result from assumption 1 that allows to stop the exploration of eliminated arms. They do not contradict the lower bound for non-stationary bandit whose scaling is in $\Omega(\sqrt{T})$ [6] as it is due to the cost of the constant exploration for the case where the best arm changes.

## 4 Non-stationary Stochastic Multi-armed Bandit with Best Arm Switches

The switching bandit problem has been proposed by Garivier et al. [6] and assumes means to be stationary between switches. In particular, the algorithm SW-UCB is built on this assumption. The setting has been extended by Allesiardo and Féraud with EXP3.R [1] to allow mean rewards to change at every time-steps. We follow this setting and consider that a best arm switch occurs when the arm with the highest mean change.

**Algorithm.** In order to allow the algorithm to choose another arm when a switch occurs, at each turn, estimators of SER3 are reseted with a probability $\varphi \in [0, 1]$ and a new task of best arm identification is started. We name this algorithm SUCCESSIVE ELIMINATION WITH RANDOMIZED ROUND-ROBIN AND RESETS (SER4).

**The sample complexity of the best arm identification problem with switches.** The **optimal policy** is the sequence of couples (optimal arm, time when the switch occurred):

$$\{(k_1^*, 1), ..., (k_N^*, T_N)\}, \qquad (6)$$

with $k_n^* \neq k_{n+1}^*$ and $\Delta_{k_n^*, k}(t) > 0$ for any $k \in [K] - \{k_n^*\}$ and any $t \in [T_n, T_{n+1})$. The optimal policy starts playing the arm $k_n^*$ at the time-step $T_n$. Time-steps $T_n$ when switches occur are unknown to the player.

**The cost of switches.** A cost associated to the number of iterations after a switch when the player does not know the optimal arm and does not sample is added to the usual sample complexity.

**Definition 1.** *Let A be an algorithm. The sample-complexity of A performing a best arms identification task for a segmentation $\{T_n\}_{n=1..N}$ of $[1:T]$, with $T_1 = 1 < T_2 < \cdots < T_N < T$, is $\sum_{n=1}^{N} \sum_{t=T_n}^{T_{n+1}-1} (s(t) + 1_{[\![k_t \neq k_n^*]\!]})$, where $s(t)$ is a binary variable equal to 1 if and only if the time-step t is used by the sampling process of A, $k_t$ is the arm identified as optimal by A at time t, $k_n^*$ is the optimal arm over the segment n and $T_{N+1} = T + 1$.*

**Performance analysis.** We now provide the performance guarantees of the SER4 algorithm, both in terms of sample complexity and in cumulative regret.

**Theorem 2.** *For $K \geq 2$, $\delta = 1/T$, $\tau_{\min} = \log \frac{K}{\delta}$ and $\varphi \in (0, 1]$, the expected sample complexity of SER4 w.r.t. the randomization of resets is upper bounded by:*

$$O\left(\frac{\varphi K}{\delta \Delta^2} \log(\frac{K}{\delta \Delta}) + \frac{N}{\varphi}\right)$$

*with a probability of at least $1 - \delta$.*

We tune $\varphi$ in order to minimize the sample complexity.

**Corollary 2.** *For $K \geq 2$, $\delta = 1/T$, $\tau_{\min} = \log \frac{K}{\delta}$, $\Delta \geq \frac{1}{KT}$ and $\varphi = \sqrt{\frac{N\delta}{K \log(\frac{K}{\delta})}}$, the expected sample complexity of SER4 w.r.t. the randomization of resets is upper bounded by:*

$$O\left(\frac{1}{\Delta^2}\sqrt{\frac{NK \log(\frac{K}{\delta})}{\delta}}\right).$$

**Remark 2:** transposing Theorem 1 for the case where $\epsilon \in [\frac{1}{KT}, 1]$ is straightforward. This allows to tune the bound by setting $\varphi = \epsilon\sqrt{(N\delta)/(K \log(TK))}$.

This result can also be transposed in bound on the expected cumulative regret. We consider that the algorithm continues to play the last arm of the set until a reset occurs.

**Corollary 3.** *For $K \geq 2$, and $\delta = 1/T$, $\tau_{\min} = \log(KT)$, $\Delta \geq \frac{1}{KT}$ and $\varphi = \sqrt{\frac{N}{TK \log(KT)}}$, the expected cumulative regret of SER4 w.r.t. the randomization of* *resets is upper bounded by:*

$$O\left(\frac{\sqrt{NTK\log(KT)}}{\Delta}\right). \qquad (7)$$

**Remark 3:** A similar dependency in $\sqrt{T}\Delta^{-1}$ appears also in EXP3.R (see Corollary 1 in [1]) and in SW-UCB (see Theorem 1 in [6]), and is standard in this type of results.

## 5 Numerical Experiments

We compare our algorithms with the state-of-the-art. For each problem, $K = 20$ and $T = 10^7$. The instantaneous gap between the optimal arm and the others is constant, $\Delta = 0.05$, i.e. the mean of the optimal arm is $\mu^*(t) = \mu(t) + \Delta$. During all experiments, probabilities of failure of SUCCESSIVE ELIMINATION (SE), SER3 and SER4 are setted to $\delta = 0.05$. Constant explorations of all algorithms of the EXP3 family are setted to $\gamma = 0.05$. Results are averaged over 50 runs. On problems 1 and 2, variances are low (in the order of $10^3$) and thus not showed. On problem 3, variances are plotted as the grey areas under the curves.

**Problem 1: Sinusoidal means.** The index of the optimal arm $k^*$ is drawn before the game and does not change. The mean of all suboptimal arm is $\mu(t) = \cos(2\pi t/K)/5 + 0.5$.

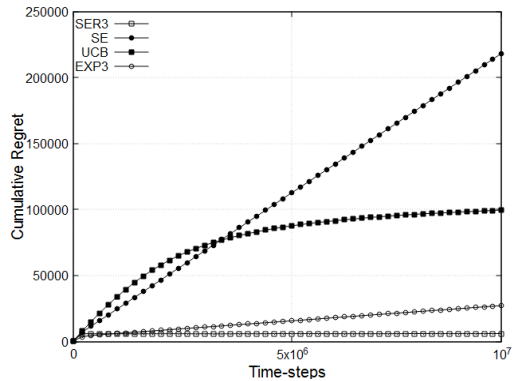

Figure 3: Cumulative regret of SER3, SE, UCB and EXP3 on the Problem 1.

This problem challenges SER3 against SE, UCB and EXP3. SER3 achieves a low cumulative regret, successfully eliminating sub-optimal arms at the beginning of the run. Contrarily, SE is tricked by the periodicity of the sinusoidal means and eliminates the optimal arm. The deterministic policy of UCB is not adapted to the non-stationarity of rewards and thus the algorithm suffers from an high regret. The unbiased estimators of EXP3 enable the algorithm to quickly converge on the best arm. However, EXP3 suffers from a linear regret due to its constant exploration until the end of the game.

**Problem 2: Decreasing means with positive gap.** The optimal arm $k^*$ does not change during the game. The mean of all suboptimal arms is $\mu(t) = 0.95 - \min(0.45, 10^{-7}t)$.

On this problem, SER3 is challenged against SE, UCB and EXP3.

SER3 achieves a low cumulative regret, successfully eliminating sub-optimal arms at the beginning of the run. Contrarily to problem 1, mean rewards evolve slowly and SUCCESSIVE ELIMINATION (SE) achieves the same level of performance than SER3. Similarly to problem 1, UCB achieves an high cumulative regret. The cumulative regret of EXP3 is low at the end of the game but would still increase linearly with time.

**Problem 3: Decreasing means with arm switches.** At every turn, the optimal arm $k^*(t)$ changes with a probability of $10^{-6}$. In expectation, there are 10 switches by run. The mean of all suboptimal arms is $\mu(t) = 0.95 - \min(0.45, 10^{-7}(t[\mod 10^6]))$.

On problem 3, SER4 is challenged against SW-UCB, EXP3.S, EXP3.R and META-EVE. The probability of reset of SER4 is $\varphi = 5^{-5}$. The size of the window of SW-UCB is $10^5$. The historic considered by EXP3.R is $H = 4 \cdot 10^5$ and the regularization parameter of EXP3.S is $\alpha = 10^{-5}$.

SER4 obtains the lowest cumulative regret, confirming the random resets approach to overcome switches of best arm. SW-UCB suffers from the same issues as UCB in previous problems and obtains a very high regret. Constant changes of mean cause META-EVE to reset very frequently and to obtain a lower regret than SW-UCB. EXP3.S and EXP3.R achieves both low regrets but EXP3.R suffers from the large size of historic needed to detect switches with a gap of $\Delta$. We can notice that the randomization of resets in SER4, while allowing to achieves the best performances on this problem, involve an highest variance. Indeed, on some runs, a reset may occur lately after a best arm switch whereas the use of windows or regularization parameters will be more consistant.

# 6 Conclusion

We studied the benefit of *random shuffling* in the design of sequential elimination bandit algorithms. We showed that the use of *random shuffling* extends their range of application to a new class of best arm identification problems involving non-stationary distributions, without requiring new parameter, while achieving the same level of guarantees than SE with stationary distributions. We introduced SER3 and extended it to the switching bandit problem with SER4 by adding a probability of restarting the best arm identification task. Up to our knowledge, we proved the first sample complexity based upper-bound for the best arm identification problem with arm switches. The upper-bound over the cumulative regret of SER4 depends only of the number $N - 1$ of arm switches, as opposed to the number of distribution changes $M - 1$ in SW-UCB ($M \geq N$ can be of order $T$ in our setting).

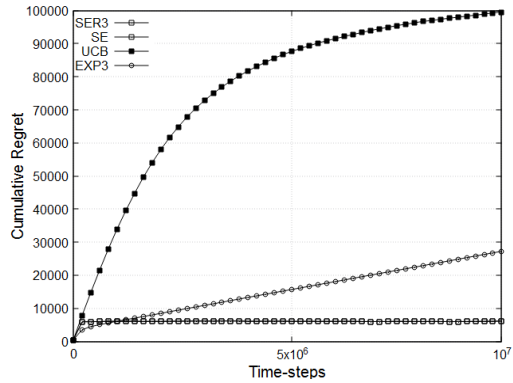

(a) Cumulative regret of SER3, SE UCB and EXP3 on the Problem 2.

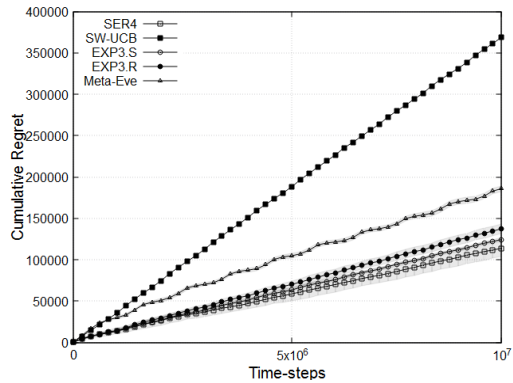

(b) Cumulative regret of SER4, SW-UCB, EXP3.S, EXP3.R and META-EVE on the Problem 3.

Figure 4

# A Technical results

## A.1 Proof of Theorem 1

The proof consists of three main steps. The first step makes explicit the conditions leading to the elimination of an arm from the set. The second step shows that the optimal arm will not be eliminated with high probability. Finally, the third step shows that a sub-optimal arm will be eliminated after at most a critical number of steps $\tau^*$, which then allows to derive an upper-bound on the sample complexity.

**Step 1. Conditions for the elimination of an arm.**

From Hoeffding's inequality, for any deterministic round-robin length $\tau$ and arm $k$ we have:

$$P\left(|\hat{\mu}_k - \mathbb{E}[\hat{\mu}_k]| \geq \epsilon_\tau\right) \leq 2\exp\left(-2\epsilon_\tau^2 \tau\right) .$$

where $\mathbb{E}$ denotes the expectation with respect to the distribution $D_y$. By setting

$$\epsilon_t = \sqrt{\frac{1}{2\tau}\log\left(\frac{4K\tau^2}{\delta}\right)}, \text{ we have:}$$

$$P\left(|\hat{\mu}_k - \mathbb{E}[\hat{\mu}_k(\tau)]| \geq \epsilon_t\right) \leq 2\exp\left(-2\sqrt{\frac{1}{2\tau}\log\left(\frac{4K\tau^2}{\delta}\right)}^2 \tau^2\right) = \frac{\delta}{2K\tau^2}.$$

Applying Hoeffding's inequality for each round-robin size $\tau \in \mathbb{N}^\star$, applying a standard union bound and using that $\sum_{\tau=1}^\infty 1/\tau^2 = \pi^2/6$, the following inequality holds simultaneously for any $\tau$ with a probability at least $1 - \frac{\delta \pi^2}{12K}$:

$$\hat{\mu}_k(\tau) - \epsilon_\tau \leq \mathbb{E}[\hat{\mu}_k] \leq \hat{\mu}_k(\tau) + \epsilon_\tau . \tag{8}$$

Let $S_i \subset \{1, \ldots, K\}$ be the set containing all the arms that are not eliminated by the algorithm at the start of the $i^{\text{th}}$ round-robin. By construction of the algorithm, an arm $k'$ remains in the set of selected arms as long as for each arm $k \in S_\tau - \{k'\}$:

$$\hat{\mu}_k(\tau) - \epsilon_\tau < \hat{\mu}_{k'}(\tau) + \epsilon_\tau \text{ and } \tau \geq \tau_{\min} \tag{9}$$

Combining (8) and the left inequality of (9), it holds on an event $\Omega$ of high probability

$$\mathbb{E}[\hat{\mu}_k(\tau)] - 2\epsilon_\tau < \mathbb{E}[\hat{\mu}_{k'}(\tau)] + 2\epsilon_\tau . \tag{10}$$

We denote $t_\tau$, the time-step where the $\tau^{\text{th}}$ round-robin starts ($t_\tau = 1 + \sum_{i=1}^{\tau-1} |S_i|$). Let us remind that $\mathbb{T}(\tau)$ is the set containing all possible realizations of $\tau$ sequences of round-robin. Each arm $k$ is played one time during each round-robin phase and thus $\tau$ observations per arm are available after $\tau^{\text{th}}$ round-robin phases. The empirical mean reward $\hat{\mu}_k(\tau)$ of each arm $k$ after the $\tau^{\text{th}}$ round-robin is:

$$\hat{\mu}_k(\tau) = \sum_{r \in \mathbb{T}(\tau)} \frac{1_{[\![r=[\tau]]\!]}}{\tau} \sum_{j=1}^{t_\tau + |S_\tau| - 1} y_k(j) 1_{[\![k=k_j]\!]} . \tag{11}$$

Decomposing the second sum in round-robin phases and taking the expectation with respect to the reward distribution $D_y$ we have:

$$\mathbb{E}_{D_y}[\hat{\mu}_k(\tau)] = \sum_{r \in \mathbb{T}(\tau)} \frac{1_{[\![r=[\tau]]\!]}}{\tau} \sum_{i=1}^{\tau} \sum_{j=t_i}^{t_i+|S_\tau|-1} \mu_k(j) 1_{[\![k=k_j]\!]} \,. \quad (12)$$

Taking the expectation of equation (12) with respect to the randomization of the round-robin we have:

$$\mathbb{E}[\hat{\mu}_k(\tau)] = \sum_{r \in \mathbb{T}(\tau)} \frac{1_{[\![r=[\tau]]\!]}}{\tau} \sum_{i=1}^{\tau} \sum_{j=t_i}^{t_i+|S_\tau|-1} \frac{\mu_k(j)}{|S_i|} \,. \quad (13)$$

Now, under the event $\Omega$ for which (10) holds for $k$ and $k'$, we deduce by using (13) that

$$\sum_{r \in \mathbb{T}(\tau)} \frac{1_{[\![r=[\tau]]\!]}}{\tau} \left( \sum_{i=1}^{\tau} \sum_{j=t_i}^{t_i+|S_\tau|-1} \frac{\mu_k(j)}{|S_i|} - \sum_{i=1}^{\tau} \sum_{j=t_i}^{t_i+|S_\tau|-1} \frac{\mu_{k'}(j)}{|S_i|} \right) < 4\epsilon_\tau \,. \quad (14)$$

Let us introduce the following mean-gap quantity

$$\Delta_{k,k'}([\tau]) = \sum_{r \in \mathbb{T}(\tau)} \frac{1_{[\![r=[\tau]]\!]}}{\tau} \left( \sum_{i=1}^{\tau} \sum_{j=t_i}^{t_i+|S_\tau|-1} \frac{\mu_k(j)}{|S_i|} - \sum_{i=1}^{\tau} \sum_{j=t_i}^{t_i+|S_\tau|-1} \frac{\mu_{k'}(j)}{|S_i|} \right) \,.$$

Replacing the value of $\epsilon_t$ in (14), it comes

$$\Delta_{k,k'}([\tau]) < 4\sqrt{\frac{1}{2\tau} \log\left(\frac{4K\tau^2}{\delta}\right)} \,,$$

$$\Delta_{k,k'}([\tau])^2 < \frac{8}{\tau} \log\left(\frac{4K\tau^2}{\delta}\right) \,. \quad (15)$$

An arm will be eliminated if (15) becomes false and if $\tau \geq \tau_{\min}$.

**Step 2. The optimal arm is not eliminated.**
For $k' = k^*$ and $k \neq k^*$, by assumption 1 ($\Delta_{k,k^*}([\tau])$ is negative after $\tau_{\min}$), (15) is always true for $\tau \geq \tau_{\min}$, implying that the optimal arm will always remain in the set with a probability of at least $1 - \frac{\delta}{K}$ for all $\tau$.

**Step 3. The elimination of sub-optimal arms.**
If the arm $k'$ still remain in the set, it will be eliminated if inequality (15) is not satisfied and if $\tau^* \geq \tau_{\min}$.

Let us consider $k = k^*$, $k' \neq k^*$, and define the quantity

$$\Delta_k([\tau]) = \sum_{r \in \mathbb{T}(\tau)} \frac{1_{[\![r=[\tau]]\!]}}{\tau} \left( \sum_{i=1}^{\tau} \sum_{j=t_i}^{t_i+|S_\tau|-1} \frac{\mu_k(j)}{|S_i|} - \sum_{i=1}^{\tau} \sum_{j=t_i}^{t_i+|S_\tau|-1} \frac{\mu_{k'}(j)}{|S_i|} \right) \,.$$

Let us also introduce for convenience the critical value

$$\tau^* = \frac{64}{\Delta_k([\tau])^2} \log\left(\frac{4K}{\delta\Delta_k([\tau])}\right).$$

Notice that $\tau^* \geq \tau_{\min}$, satisfying one of the two conditions needed to eliminate an arm.

We introduce the following quantity

$$C(t) \stackrel{\text{def}}{=} \frac{8}{\tau} \log\left(\frac{4K\tau^2}{\delta}\right).$$

For $\tau = \tau^*$, we derive the following bound

$$\begin{aligned}
C(\tau^*) &= \frac{\Delta_k([\tau])^2}{8\log\frac{4K}{\delta\Delta_k([\tau])}} \left(\log\frac{4K}{\delta} + 2\log\frac{64K^2}{\Delta_k([\tau])^2} + 2\log\log\frac{4K}{\delta\Delta_k([\tau])}\right), \\
&= \frac{\Delta_k([\tau])^2}{8\log\frac{4K}{\delta\Delta_k([\tau])}} \left(\log\frac{4K}{\delta} - 4\log\Delta_k([\tau]) + 12\log 2 + 2\log\log\frac{4K}{\delta\Delta_k([\tau])}\right), \\
&\leq \frac{\Delta_k([\tau])^2}{8\log\frac{4K}{\delta\Delta_k([\tau])}} \left(4\log\frac{4K}{\delta\Delta_k([\tau])} + 12\log 2 + 2\log\log\frac{4K}{\delta\Delta_k([\tau])}\right).
\end{aligned}$$

We remark that for $X > 13$ we have

$$12\log 2 + 2\log\log X < 4\log X.$$

Hence, provided that for $K \geq 2$, $\delta \in (0, 0.5]$ and $\Delta_k([\tau]) > 0$, we have $\frac{4K}{\delta\Delta_k([\tau])} > 13$ and

$$\begin{aligned}
C(\tau^*) &\leq \frac{\Delta_k([\tau])^2}{8\log\frac{4K}{\delta\Delta_k([\tau])}} \left(8\log\frac{4K}{\delta\Delta_k([\tau])}\right) \\
&\leq \Delta_k([\tau])^2. \quad (16)
\end{aligned}$$

As $C(\tau^*)$ is strictly decreasing with regard to $t$, (16) is true for all $\tau > \tau^*$, invalidating (15) and involving the elimination of the suboptimal arms $k$ with a probability at least $1 - \delta/K$.

We conclude the proof by summing over all the arms, taking the union bound and lower-bounding all $\Delta_k([\tau])$ by

$$\Delta = \min_{[\tau]\in\mathbb{T}(\tau), k} \sum_{r\in\mathbb{T}(\tau)} \frac{\mathbb{1}_{[\![r=[\tau]]\!]}}{\tau} \left(\sum_{i=1}^{\tau}\sum_{j=t_i}^{t_i+|S_\tau|-1} \frac{\mu_k(j)}{|S_i|} - \sum_{i=1}^{\tau}\sum_{j=t_i}^{t_i+|S_\tau|-1} \frac{\mu_{k'}(j)}{|S_i|}\right). \tag{17}$$

□

## A.2 Proof of Corollary 1

We first provide the proof of the distribution dependent upper-bound.

The cumulated pseudo regret of the algorithm is:

$$R(T) = \sum_{k \neq k^*} \sum_{i=1}^{\tau} \sum_{t=t_i}^{t_i+|S_i|-1} \Delta_{k^*,k}(t) 1_{[\![k=k_t]\!]} . \tag{18}$$

Taking in each round-robin the expectation of the corresponding random variable $k_t$ with respect to the randomization of the round-robin (denoted by $\mathbb{E}_{k_t}$), it comes:

$$\mathbb{E}[R(T)] = \mathbb{E}\left[\sum_{k \neq k^*} \sum_{i=1}^{\tau} \sum_{t=t_i}^{t_i+|S_i|-1} \mathbb{E}_{k_t}[\Delta_{k^*,k}(t) 1_{[\![k=k_t]\!]}]\right]$$

$$= \mathbb{E}\left[\sum_{k \neq k^*} \sum_{i=1}^{\tau} \sum_{t=t_i}^{t_i+|S_i|-1} \frac{\Delta_{k^*,k}(t)}{|S_i|}\right] .$$

$$\mathbb{E}[R(T)] = \mathbb{E}\left[\sum_{k \neq k^*} \tau \underbrace{\frac{1}{\tau} \sum_{i=1}^{\tau} \sum_{t=t_i}^{t_i+|S_i|-1} \frac{\Delta_{k^*,k}(t)}{|S_i|}}_{\Delta_k^*}\right] = \mathbb{E}\left[\sum_{k \neq k^*} \tau \Delta_k^*\right] . \tag{19}$$

The penultimate step of the proof of Theorem 1 allows us to upper-bound $\tau$ with the previously introduced critical value $\tau^*$ on an event of high probability $1 - \delta$, while the cumulative regret is controlled by the trivial upper bound $T$ on the complementary event of probability not higher than $\delta$, leading to:

$$\mathbb{E}[R(T)] \leq \sum_{k \neq k^*} \frac{64}{\Delta_k^2} \log\left(\frac{4K}{\delta \Delta_k}\right) \Delta_k + \delta T . \tag{20}$$

We conclude the proof of the distribution dependent upper-bound by setting $\delta = 1/T$ and :

$$\mathbb{E}[R(T)] = O\left(\frac{K-1}{\Delta} \log(\frac{KT}{\Delta})\right) , \tag{21}$$

with $\Delta = \min_{[\tau],k} \frac{1}{\tau} \sum_{i=1}^{\tau} \sum_{t=t_i}^{t_i+|S_i|-1} \frac{\Delta_{k^*,k}(t)}{|S_i|}$.

We now upper-bound the regret in the worst case in order to derive a distribution independent bound. To this end, we consider a sequence that ensures that, with high probability, no suboptimal arm is eliminated by the algorithm at the end of the $T$ rounds, while maximizing the instantaneous regret. According to (10) an arm is not eliminated as long as

$$\mathbb{E}[\hat{\mu}_k(\tau)] - \mathbb{E}[\hat{\mu}_{k'}(\tau)] < 4\epsilon_\tau . \tag{22}$$

By injecting (22) in (19) and replacing $\epsilon_\tau$ by its value $\sqrt{\frac{2}{\tau} \log\left(\frac{4K\tau^2}{\delta}\right)}$ we obtain:

$$\mathbb{E}[R(T)] < \sum_{k \neq k^*} \tau 4 \sqrt{\frac{2}{\tau} \log\left(\frac{4K\tau^2}{\delta}\right)} + \delta T . \tag{23}$$

The non-elimination of sub-optimal arms involves $\tau = \frac{T}{K}$ and by setting $\delta = \frac{1}{T}$ we obtain the distribution independent upper-bound:

$$\mathbb{E}[R(T)] < (K-1)\frac{T}{K}4\sqrt{\frac{K}{T}\log(\frac{4T^3}{K})} + 1,\qquad(24)$$

$$\mathbb{E}[R(T)] = O\left(\sqrt{TK\log\frac{T}{K}}\right).\qquad(25)$$

□

### A.3 Proof of Theorem 2

In order to prove Theorem 2, we consider the following quantities:

- The expected number of times when the estimators are reseted: $N_{\text{reset}} = \varphi T$.

- The sample complexity needed to find the best arm between each reset is $S_{\text{SER3}} = O\left(\frac{K}{\Delta^2}\log(\frac{K}{\delta\Delta})\right)$.

- The time before a reset, that follows a negative binomial distribution of parameters $r = 1$ and $p = 1 - \varphi$. Its expectation is upper-bounded by $1/\varphi$.

- The number of arm switches: $N - 1$.

The sample complexity of SER4 is the total number of time-steps spent sampling an arm added to the time between each switch and reset.

Taking the expectation with respect to the randomization of resets, we obtain an upper-bound on the expected number of suboptimal plays given by

$$O\left(\frac{\varphi TK}{\Delta^2}\log\left(\frac{K}{\delta\Delta}\right) + \frac{N}{\varphi}\right).\qquad(26)$$

The first term is the expectation of the total number of time-steps required by the algorithm in order to find the best arms at its initialization and then after each reset of the algorithm. The second term is the expected total number of steps lost by the algorithm when not resetting the algorithm after the $N-1$ arm switches.

We obtain the final statement of the Theorem by setting $T = \frac{1}{\delta}$. □

### A.4 Proof of Corollary 3

Converting Corollary 2 into a distribution dependent upper-bound on the cumulative regret is straightforward by setting $\delta = \frac{1}{T}$, replacing the sample complexity in the proof of Theorem 2 by the cumulative regret and using the upper-bound of Corollary 1.

$$\mathbb{E}[R(T)] = O\left(\frac{\varphi TK}{\Delta}\log\left(\frac{KT}{\Delta}\right) + \frac{N}{\varphi}\right).\qquad(27)$$

Setting $\varphi = \sqrt{\frac{N}{TK\log(KT)}}$ and assuming $\Delta \geq \frac{1}{KT}$ we obtain the final statement of the theorem:

$$\mathbb{E}[R(T)] = O\left(\frac{\sqrt{NTK\log(KT)}}{\Delta}\right). \tag{28}$$

We also derive below a distribution independent upper-bound. We introduce some notations, $N_{\text{reset}}$ is the number of resets, $\tau_i^{\text{reset}}$ is the number of round-robin phases between the $i^{\text{th}}$ and the $(i+1)^{\text{th}}$ resets and $L_n$ is the number of timesteps before a reset after the $n^{\text{th}}$ arm switch.

When the resets are fixed, the expected cumulative regret is:

$$\mathbb{E}[R(T)] < \mathbb{E}\left[\sum_{i=1}^{N_{\text{reset}}+1}(K-1)\tau_i^{\text{reset}}4\sqrt{\frac{2}{\tau_i^{\text{reset}}}\log(\frac{4(\tau_i^{\text{reset}})^2}{\delta})} + \sum_{n=1}^{N}L_n + \delta T\right], \tag{29}$$

$$\mathbb{E}[R(T)] < \mathbb{E}\left[\sum_{i=1}^{N_{\text{reset}}+1}\underbrace{(K-1)4\sqrt{2\tau_i^{\text{reset}}\log(\frac{4(\tau_i^{\text{reset}})^2}{\delta})}}_{f(\tau_i^{\text{reset}})}\right] + \mathbb{E}\left[\sum_{n=1}^{N}L_n\right] + \delta T. \tag{30}$$

At this point, we note that $\{\tau_i^{\text{reset}}\}_i$ is an i.i.d sequence of random variables and that $N_{\text{reset}}$ is a random stopping time with respect to this sequence. Moreover, $f$ is a concave function. We can thus apply Wald's equation followed by Jensen's inequality and deduce that

$$\mathbb{E}[\sum_{i=1}^{N_{\text{reset}}+1}f(\tau_i^{\text{reset}})] \leq \mathbb{E}[N_{\text{reset}}+1]\mathbb{E}[f(\tau_1^{\text{reset}})]$$

$$\leq \mathbb{E}[N_{\text{reset}}+1]f(\mathbb{E}[\tau_1^{\text{reset}}]).$$

We upper-bound $\log(\frac{4(\tau_i^{\text{reset}})^2}{\delta})$ by $\log(\frac{4T^2}{\delta K^2})$ and set $\delta = \frac{1}{T}$. As $\mathbb{E}[N_{\text{reset}}] = \varphi T$, $\mathbb{E}[\tau_1^{\text{reset}}] = \frac{1}{\varphi K}$ and $\mathbb{E}[L_n] \leq \frac{1}{\varphi}$, we have

$$\mathbb{E}[R(T)] < 4(\varphi T + 1)\sqrt{\frac{2}{\varphi}K\log\left(\frac{4T^3}{K^2}\right)} + \frac{N}{\varphi} + 1. \tag{31}$$

The previous equation makes appear a trade-off in $\varphi$, and we set $\varphi = \frac{\sqrt{N}}{T^{2/3}}$.
Finally we have shown that

$$\mathbb{E}[R(T)] = O\left(T^{2/3}\sqrt{NK\log\frac{T}{K}}\right). \tag{32}$$

$\square$